# Lazy Factored Inference for Functional Probabilistic Programming

**Avi Pfeffer, Brian Ruttenberg, Amy Sliva, Michael Howard, and Glenn Takata**
Charles River Analytics
625 Mt. Auburn St
Cambridge, MA, 02138
{apfeffer, bruttenberg, asliva, mhoward, gtakata}@cra.com

## Abstract

Probabilistic programming provides the means to represent and reason about complex probabilistic models using programming language constructs. Even simple probabilistic programs can produce models with infinitely many variables. Factored inference algorithms are widely used for probabilistic graphical models, but cannot be applied to these programs because all the variables and factors have to be enumerated. In this paper, we present a new inference framework, lazy factored inference (LFI), that enables factored algorithms to be used for models with infinitely many variables. LFI expands the model to a bounded depth and uses the structure of the program to precisely quantify the effect of the unexpanded part of the model, producing lower and upper bounds to the probability of the query.

## 1. INTRODUCTION

Probabilistic models are growing in their richness and diversity. One of the challenges to using probabilistic models is the need to create representations and reasoning algorithms. Probabilistic programming (PP) (Koller, McAllester, and Pfeffer, 1997) addresses these challenges by providing expressive languages to represent models using programming language constructs and inference algorithms that apply to models written in the languages.

One of the biggest challenges in PP inference is that even compact programs can result in models with a very large or infinite number of variables. Currently, the typical method for performing inference in PP systems based on functional programming is to use Metropolis-Hastings (MH) (Metropolis, Rosenbluth, and Rosenbluth, 1953; Hastings, 1970), which has become a standard algorithm in languages such as BLOG (Milch et al., 2005), Church (Goodman et al., 2008), and Figaro (Pfeffer, 2012). Unfortunately, MH is extremely hard to understand and requires significant expertise to achieve convergence in many applications.

Factored algorithms, such as junction tree (Lauritzen and Spiegelhalter, 1988), variable elimination (VE) (Zhang and Poole, 1994; Dechter, 1999) and belief propagation (BP) (Pearl, 1988; McEliece, Mackay, and Cheng, 1998), are widely used inference algorithms and are generally preferred to MH. In the 2010 UAI Approximate Inference Competition, many of the entrants used factored algorithms, while none used MH. However, current factored algorithms require enumerating all the variables in the model and creating factors for them, which is infeasible for models with a very large number of variables, and impossible if there are infinitely many variables. Indeed, Infer.NET (Winn, 2008) has achieved excellent results on real-world inference tasks (Herbrich, Minka, and Graepel, 2006) using expectation propagation (Minka, 2001), a factored algorithm, at the cost of severely restricting the expressivity of the language to avoid recursion and eliminate infinite models.

Just as factored algorithms have been instrumental in the success of probabilistic graphical models in general, making factored inference work for PP is essential to its eventual success. In this paper we describe an inference framework—*lazy factored inference (LFI)*—that achieves this goal, making factored algorithms applicable to models with very many or infinitely many variables. LFI expands a potentially infinite model up to a bounded depth and characterizes precisely the effect of the unexpanded part of the model on the probability of the query. This characterization can be done using standard factored inference algorithms as a subroutine, with the addition of preprocessing and postprocessing steps. The result of LFI is lower and upper bounds on the probability of the query. By iterative expansion of the model to increasing depths, we obtain an anytime algorithm that can produce progressively tighter bounds. LFI is a general inference framework that works well for PP because PP languages typically have the necessary constructs to guide the lazy expansion.

After providing a running example that will be used to illustrate LFI, we present the basic intuition and technical details of LFI. Next, we present theoretical results and analysis of the LFI approach showing the correctness of the bounds produced by the algorithm. We then describe an implementation of two lazy factored algorithms—VE and BP—in the open source Figaro PP language (Pfeffer, 2012) and present experimental results on reasoning with infinite hidden Markov models and probabilistic context-free grammars, which would otherwise be intractable using standard factored algorithms. We then discuss connections to and differences from work on logical probabilistic programming languages, as well as other related work in probabilistic reasoning and lazy evaluation.

## 2. RANDOM LISTS EXAMPLE

As a simple running example, we use a model that generates random lists of unbounded length. Each list consists of the symbol 'a or the symbol 'b at each index. Lists are created by a generator function that grows the list one symbol at a time. At each step, the generator terminates with probability 0.5, adds an 'a with probability 0.3, or adds a 'b with probability 0.2. We can query the list for certain properties, such as whether it contains a 'b.

This list generator and the containment queries can both be defined in Figaro, a PP language embedded in Scala. The following code defines the list using the Figaro `Element` construct, which represents random variables. A random list is either Empty, or it is a Cons, where the head is an `Element[Symbol]` and the tail is an `Element[L]` (i.e., a random list).

```
abstract class L
case object Empty extends L
case class Cons(head: Element[Symbol],
        tail: Element[L]) extends L
```

We now define the random list generator function. First, it uses `Flip(0.5)` to generate a random Boolean that is true with probability 0.5. If the Boolean is true, it produces `Empty`. Otherwise, it produces a `Cons` where the head is 'a with probability 0.6 and 'b with probability 0.4, and the tail is a recursive call to `generate()`. Sampling from this function could generate unbounded lists or a model with infinitely many variables.

```
def generate(): Element[L] = {
 Apply(Flip(0.5), (b: Boolean) =>
  if (b) Empty
  else Cons(Select(0.6 -> 'a, 0.4 -> 'b),
        generate()))}
```

Suppose we want to know whether this list contains a particular symbol. We can define a `contains` predicate over the `target` symbol and the random list `el` we are checking.

```
def contains(target: Symbol,
 el: Element[L]): Element[Boolean] = {
 Chain(el, (l: L) => {
   l match {
    case Empty => Constant(false)
    case Cons(head, tail) =>
     If(head === target,
        Constant(true),
        contains(target, tail))}})}
```

The result of the `contains` predicate is a random variable denoted by the type `Element[Boolean]`. `contains` uses `Chain`, a Figaro construct that chains random processes together through two arguments: an Element (random variable) and a function that takes a value of the `Element` and produces another `Element`. In the case of contains, the `Element` argument is the random list `el`. The function argument takes a particular value of `el`, which is a list `l`, and returns an `Element[Boolean]`. If `l` is `Empty`, the function returns `Constant(false)`. Otherwise, if the value of head is equal to the `target`, it returns `Constant(true)`, otherwise it recursively calls contains on the tail.

Using this model, we want to query about the contents of a random list. The code below first generates a random list and then creates random Booleans indicating whether `el` contains the symbols 'a or 'b. Next, it sets the observation that the random List contains 'a. Given this evidence, we want to determine the probability that the list also contains 'b. Although the answer can be determined analytically in this simple example, a general algorithmic solution would need to sum over infinitely many sequences of unbounded length, motivating the need for a lazy solution. The final line creates a lazy version of VE capable of solving this otherwise intractable query.

```
val el = generate()
val ca = contains('a, el)
val cb = contains('b, el)
ca.observe(true)
val alg = new LazyVariableElimination(cb)
```

## 3. THE LFI ALGORITHM FOR PROBABILISTIC PROGRAMMING

So, how do we make a factored algorithm work without enumerating the infinitely many variables in the model? The main intuition is that variables that are far from the query and evidence have little impact on the query. However, it is not just the distance from the query and evidence that matters; it is the fact that other variables need to take on particular values to make these faraway variables relevant. In other words, those variables may be contextually independent of the query given the values of expanded variables.

LFI expands the model up to a bounded depth, exploring only *relevant* parts of the model. For example, from a partial expansion of our random list model to the first $n$ elements of a list $l$, we can compute:

- $p_1$ = P($l$ has length $\leq n$ and does not contain 'b)
- $p_2$ = P($l$ contains 'b in the first $n$ elements)
- $p_3$ = P($l$ has length $> n$ and does not contain 'b in the first $n$ elements)

In the first case, the query for whether the list contains 'b is definitely false, in the second case it is definitely true, and in the third case the query is not yet determined. So ($p_2$, $p_2 + p_3$) are lower and upper bounds on the probability that the list contains 'b. When we have evidence that the list contains 'a, we get more cases, but the principle is similar.

Of course, since we are only partially exploring the model along relevant paths, we cannot guarantee that all unexplored portions are irrelevant to the query. To represent the unexplored probability mass in LFI, we extend the range of values a variable can take.

**Definition 1 (Extended Range).** *A variable with an* extended range *can take a regular value, or it can take the special value * (pronounced "star").*

For example, the possible extended values of a Boolean are { false, true, * }. Intuitively, * stands for "unknown result of the rest of the computation," and the probability associated with * represents the amount of probability mass resulting from the unexplored part. If we quantify this, we know how much remaining probability mass could be added to each of the regular values. As will be discussed later, by computing sums and products involving extended values in the ordinary way, we can keep track of this probability mass.

The above concepts can be formalized into a LFI algorithm consisting of four steps:

1. Expand the model to the desired depth and compute the extended ranges of relevant elements
2. Produce factors for the relevant elements
3. Apply a factored inference algorithm to the factors
4. Finalize the result to produce bounds on the query

## 4. THE STEPS OF LFI

We now provide details on the four steps of the LFI algorithm for PP and its implementation using Figaro.

### 4.1 STEP 1: EXPAND THE MODEL

The first step of the LFI algorithm is to expand the model, beginning with the query and evidence, up to a depth $d$. This step must determine which variables are relevant when the model is expanded to this depth and the range of each relevant variable, which is a set of extended values (possibly including *). We present two approaches to the expansion, a basic algorithm (Section 4.1.1) suitable for simple queries, and a backtracking version (Section 4.1.2) that can be used to compute more complex queries and evidence. Section 4.1.3 specifically addresses lazy expansion of evidence.

We explain these algorithms using Figaro constructs, but they are all generalizable to other functional PP languages. Recall from Section 2 that in Figaro a random variable is represented by an Element. Some elements are atomic, meaning they do not depend on any arguments (e.g., Select(0.6 -> 'a, 0.4 -> 'b) is the probabilistic model that produces 'a with probability 0.6 and 'b with probability 0.4). An element can also consist of the more complex Chain structure for chaining random processes together (see Section 2). As we will see in Section 4.2, the Chain construct helps to control and limit the impact of the unexplored part of the computation on the query. Most functional probabilistic programming languages have a structure similar to Chain that can be used in this manner. Finally, an element can have the form Apply(*arguments*, *function*), in which the arguments are elements, and the Apply element corresponds to the random variable produced by applying the deterministic *function* to the arguments. Since Figaro constructs can in general be expressed in terms of atomic elements, Chain, and Apply, it suffices to define the algorithm for these element classes.

#### 4.1.1 Basic expansion algorithm

The basic expansion algorithm begins with a list of relevant elements consisting of the query and evidence, represented as Figaro elements, and proceeds recursively to depth $d$ as follows.

For a relevant element $E$:

1. If $d < 0$, return { * } for the range of $E$
2. If $E$ is atomic, return its known range of regular values.
3. If $E$ is a Chain($X, F$), where $X$ is an element and $F$ is a function that maps a value of $X$ to another element:
    a. Expand X to depth $d$ - 1.
    b. For each regular value $x$ in the range of $X$:
        i. Compute $Y = F(x)$.
        ii. Expand Y to depth $d$ - 1.
        iii. Each value, regular or *, in the range of $Y$ is added to the range of $E$.
    c. If the range of $X$ includes *, the range of $E$ also includes *.

4. If *E* is `Apply(X,F)`, where **X** is a sequence of argument elements and *F* is a deterministic function of values of **X**:
    a. Expand each *X* in **X** to depth *d* - 1.
    b. For each combination **x** of regular values of **X**, the range of *E* contains *F*(**x**).
    c. If any argument in **X** contains * in its range, the range of `E` also contains *.

All the elements that are expanded in this way, including those that are expanded to a depth of -1 and so have the range { * }, are *relevant*. At the end of this step, we create a variable for each such element; its range is the computed range of extended values of the element. These variables are later used to produce factors for the inference algorithms.

Figure 1 shows an example of the basic expansion algorithm for our random list example. Each node in the graph in Figure 1 corresponds to an element whose values are to be computed, and the shaded box beneath shows the resulting values. The numbers in parentheses to the left of the elements and the resulting values indicates the order in which the elements were expanded and their values were determined. The small superscript number to the right of the element represents the depth to which the element is expanded.

In this example, we want to determine which variables are relevant to our query—whether the list `el` contains the symbol `'b`—by looking to a target depth of *d* = 3. The first step is to expand the top-level query, `contains('b, el)` to *d* = 3. This query is a compound element, so will expand its arguments in Step (2) to depth *d*-1 = 2. The algorithm first looks at the value of `el`, which is defined by a call to `generate()`, and expands `generate()` to *d* = 2. Again, we have a compound element, so the arguments of the `generate()` element are expanded to depth *d* = 1. Step (3) first looks at `Flip(0.5)`, which immediately produces the values {F, T} in Step (4). There are then two possible outcomes, depending on the value of the `Flip`: `Empty`, and `Cons(Select(0.6 -> 'a, 0.4 -> 'b), generate())`, which produce the value sets {`Empty`} and {`Cons`} respectively in Steps (5)-(8) as they are expanded and their ranges computed. Note that even though `Cons` contains two random elements, `Cons` itself is just a value. So, in Step (9), we determine that the possible values of `generate()` are {`Empty, Cons`}. If `generate()` is `Empty`, the top level query is `Constant(false)`, whose value set is {F} (Steps (10)-(11)), so F will become a possible value for the top level query.

In this depth 3 expansion, so far we have found the case where the generated list is empty. Otherwise, the top level query is the result of `If(head === 'b, Constant(true),contains(target, tail))`. In Step (13), we expand this compound element starting with expanding `head`, which we get out of the previously computed `Cons`. The range of values for `head` are {`'a,`

`'b`} (Step (14)), so the values of `head === 'b` are {F, T}. Since the test for `'b` in the `head` could be either T or F, we expand both consequences. In the first consequence (Step (17)), the head of the list is `'b`, so we have found a case where the top level query has value T. In the other consequence, we have a recursive call to `contains('b, tail)` at depth *d* = 1 in Step 19. This results in an expansion of `generate()` at depth 0, which in turn results in expansion of `Flip(0.5)` at *d* = -1 in Step (21). Since the depth is negative, we immediately get the result {*} for the range of `Flip(0.5)`. Since the `Flip` has no regular values, we do not expand either of the two outcomes `Empty` or `Cons`. Instead, we immediately return the value set {*} for `generate()`, and in turn for `contains('b, tail)`. This corresponds to a possible value of * for the top level query. In the end in Step (26), we get the value set {F, T,*} for the top level query.

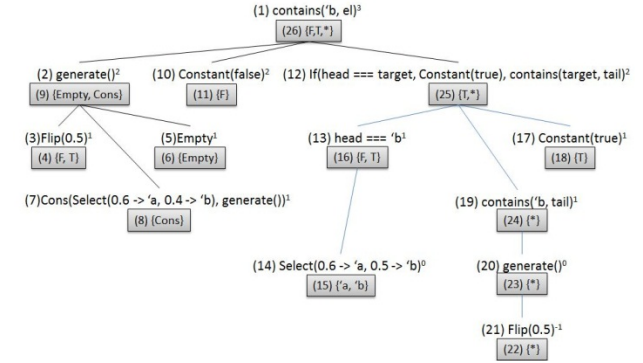

Figure 1: Basic LFI expansion on a random list.

### 4.1.2 Backtracking Expansion

The above algorithm is sufficient when we are only expanding a single query with no evidence and when the expansion forms a tree such that no element occurs in more than one path. However, if the same element is used both by the query and some evidence, or is reachable from the query by more than one path, this basic expansion algorithm encounters a subtle problem where it may compute inconsistent ranges for the same elements.

Suppose we have a query element *X* and an evidence element *Y*, and the target depth is 1. Suppose also that *Y* is an argument of an argument of *X*. If we expand *X* first, we will eventually expand *Y* to depth -1, resulting in a range of { * }. However, because *Y* is an evidence element, we will eventually expand it to depth 1, resulting in a different range. The computed range of *Y* will be incompatible with the range of *X*, which can cause trouble for factored computation later on.

One possible solution is to stipulate in advance that whenever an important (query or evidence) element is encountered, it is always expanded to the maximum desired depth *d*. However, this does not completely solve the problem, because X and Y might both depend, at different depths, on some other element Z that is not a query or evidence element.

Our solution uses backtracking to keep track of dependencies at various depths and adjust previous computations once new dependencies are revealed by the expansion algorithm. Failure to make this optimization can lead to exponential blowup as the same elements get recursively expanded again and again. Consider a case where X and Y both depend on an element Z. Suppose Y is expanded first, resulting in Z being expanded to some depth $d_1$. After Z has been expanded, we record a back pointer from Z to Y. When X is later expanded, it will result in a request to expand Z to depth $d_2$. If $d_2 \leq d_1$, we have already computed an equal or better set of values for Z, so we do not expand Z again. If, however $d_2 > d_1$, we need to expand Z to a greater depth. After doing so and computing a new set of values for Z, we know from the back pointer that Z was previously expanded from Y to a lesser depth than $d_2$, so Y might use an inconsistent set of values of Z. Therefore, we backtrack and re-expand Y. We will also have back pointers from Y so we can re-expand other elements that depend on Y.

Using backtracking, we can ensure that the last time the values of an element are computed in the basic expansion algorithm occurs after the last time values have been computed for all elements on which it depends.

**Proposition 1.** *For all elements Y that have been expanded by the LFI expansion algorithm with backtracking, the last expansion for Y occurs after the last expansion of all elements on which it depends.*

Please see the supplement for all proofs. This fact ensures that the value sets will be consistent.

### 4.1.3 Lazily Expanding Evidence

There is an additional optimization we can make to the expansion phase of LFI. Consider a large model with many evidence elements and a single query. Implementing the above expansion algorithm will require us to expand all the evidence variables regardless of their distance from the query, resulting in a large number of elements. However, as with irrelevant parts of the model that are represented by *, distant evidence may not be relevant to the query (i.e., there will be no path from the evidence variables to a query variable within depth d of the query). Ideally, we will only expand evidence that is close to the query and can actually contribute to the probability bounds computation.

We can accomplish this by modifying the basic expansion algorithm to lazily expand in multiple iterations, beginning with only the query elements Q.

1. Set `ExpandList = Q` with depth d
2. For each element E in `ExpandList`, expand E to specified depth d as described in Section 4.1.1
3. For each iteration where $d \geq 0$
   a. Identify all elements X that use the current element E and have not be expanded to d - 1
   b. If X is an evidence element, then add to `ExpandList` with depth d - 1
   c. Recursively expand X until $d < 0$
4. Continue until `ExpandList = Empty`

After this process has completed, we guarantee that all elements relevant to the query within a distance of d have been expanded.

**Theorem 1.** *Let **Q** denote a set of query variables and **E** a set of evidence variables with known values in a probabilistic graphical model **G**. Lazy expansion of **G** to depth d will expand all variables relevant to **Q** and **E** within depth d of **Q**.*

## 4.2 STEP 2: PRODUCE FACTORS FOR THE RELEVANT ELEMENTS

Once the model has been lazily expanded to the desired depth to identify the relevant elements and their possible values, the next step of the LFI algorithm is to produce factors for these elements so they can be used in a factored inference algorithm. However, while the lazy expansion has produced a finite set of elements to be used in factor generation, these factors must still account for variables with extended ranges. Using constructs from PP, we can produce factors even over the value * representing the unexplored probability mass.

The following algorithm is again explained using Figaro constructs, but is generalizable to other PP languages.

Figaro already contains an algorithm for producing factors for a finite set of elements whose corresponding variables have ordinary (not extended) ranges. Producing factors for elements whose variables have extended ranges extends this procedure in a straightforward way.

In general, there are two kinds of factors produced by Figaro. The first encodes the relationship between an element and its arguments resulting from the definition of the element's generative model. The second encodes conditions and constraints on a variable.

For the first kind of factor for an element E:

1. If E is atomic, its factor is the usual factor over its regular values.
2. If E is `Apply(X,F)`, then the factor assigns a probability to each assignment **x** to the arguments and y to the result, as follows:
   a. If none of the arguments are *, and $y = F(\mathbf{x})$, the probability is 1.
   b. If any of the arguments is *, and y = *, the probability is 1.
   c. Otherwise, the probability is 0.

3. If $E$ is Chain($X$, $F$), then we build off a technique in Figaro for constructing factors for a chain. Since every value of $X$ results in a different element, a naïve factor would include a variable for each such element, potentially resulting in extremely large factors. Instead, many three variable factors are constructed. For each regular value $x$ of $X$, we construct a factor $\varphi_x$ over $X$, the specific element $Y = F(x)$ for some value $x$ of $X$, and $E$. Without extended values, these factors are defined such that their product equals the single naïve factor. We extend this construction to extended values as follows:
    a. For each regular value $x$ of $X$, we define a factor $\varphi_x$ that specifies a probability for each value $x'$ of $X$, $y$ of $F(x)$, and $e$ of $E$, as follows:
        i. If $x' \neq x$, the probability is 1. This is a "don't care" case.
        ii. If $x' = x$ and $e = y$, the probability is 1. This also applies if $e = y = *$.
        iii. Otherwise the probability is 0.
    b. We also create a binary factor $\varphi_*$ that specifies a probability for each value $x$ of $X$ and $e$ of $E$, as follows:
        i. If $x \neq *$, the probability is 1 (don't care).
        ii. If $x = *$ and $e = *$, the probability is 1.
        iii. Otherwise the probability is 0.

To see how this construction for chains helps control the effect of *, consider the following element from our random list example:

```
If(head == target, Constant(true),
contains(target, tail))
```

If is actually syntactic sugar for Chain, in which the first argument is the test, and the function maps the result of the test to the appropriate consequence. Here, if the test is true (i.e., the value of head is equal to the target symbol), only the then clause Constant(true) is relevant, so the factor $\varphi_{\text{true}}$ will not include the variable for the else clause, while the factor $\varphi_{\text{false}}$ will have a don't care case. Therefore, even if the value of the else clause is *, the value true for the entire If expression will have probability 1 in each factor. This is the essential insight that prevents * contaminating the entire computation.

The constraint factor corresponds to conditions or constraints on the element that are either satisfied or not satisfied. To produce a factor for an element $E$ and condition C:

1. If $E$ has a regular value, we can determine if C is satisfied and compute an entry of 0 or 1 as usual.
2. If the value of $E$ is *, we do not know if C would be satisfied by the eventual value * would resolve to if we expanded it fully, so we create bounds of [0, 1] on the entry.

Factors representing soft constraints, which are functions from the value of a variable to a real number, are similar. Bounds must be specified on the value of the constraint. Bounds of [0, 1] are the default, but different or more precise bounds can be provided as necessary.

Using these modifications to Figaro's factor generation algorithm to account for unexpanded parts of the computation represented by *, Step 2 will produce a set of factors over variables with extended ranges, which can subsequently be used in a factored inference algorithm in PP. Only factors for relevant variables within the desired depth will be produced.

### 4.3 STEP 3: APPLY A FACTORED ALGORITHM

Using the factors produced by Step 2, we can now determine an answer to the query, which is defined as a sum-of-products expression over these factors. The goal is to reduce this sum-of-products expression to a single factor over the query variables. Factored algorithms such as VE and BP produce solutions or approximations to this factor.

For LFI, standard factored inference algorithms can be applied with no modification; however, they are only computed over factors representative of the relevant parts of the computation for answering the query to the desired depth. The standard algorithm is called once using the lower bounds and once using the upper bounds specified in the factors.

### 4.4 STEP 4: FINALIZE THE RESULT

By applying a factored inference algorithm in the previous step, we acquire two factors over the query, one each for the lower and upper bounds. These factors will, in general, be unnormalized, and * might have positive probability mass. In this finalization step, we need to normalize the results and absorb the probability mass of * into the regular values.

Let the unnormalized lower bound of value $i$ (regular or *) of the query be $l_i$ and let the unnormalized upper bound be $u_i$. Standard normalization takes a set of unnormalized probabilities $q_i$, computes their sum $Z = \Sigma q_i$, and then computes $p_i = q_i / Z$ to obtain the normalized probabilities. In our case, $U = \Sigma u_i$ is an upper bound on the normalizing factor. Therefore $L_i = l_i / U$ is a lower bound on the normalized probability of value $i$. Meanwhile, for a regular value $j$ of the query, any probability assigned to the regular value $i \neq j$ cannot be assigned to $j$, so $1 - \Sigma_{\text{regular } i \neq j} L_i$ is an upper bound on the probability of $j$. Since any probability mass associated with * will not be subtracted in this upper bound, that probability mass is absorbed into the upper bounds of the regular values.

## 5. ANALYSIS

Our main result is that the process of lazily expanding the program to increasing depths results in increasingly better bounds on the probability distribution over the query. Our analysis assumes there is a single variable, and in fact multiple query variables can break the result if query variables only become connected after some depth has been expanded. If multiple query variables are desired, that can easily be achieved by defining a single variable to be a tuple of the query variables, and making that the query variable instead.

Our result assumes that all evidence variables have already been included before the bounds start to converge. If new evidence variables are introduced after a certain depth, they might change the query distribution. In many applications, such as the probabilistic context free grammar example we present later, this is not a problem as the evidence is reached at a shallow depth. Our result assumes that the factored algorithm used to compute the bounds is exact. For an approximate algorithm like BP, we cannot provide the same guarantees.

Our main result is as follows:

**Theorem 2.** *Let $Q$ be a query variable, $E$ a set of evidence variables, and $q$ a regular value of $Q$. Assume that expanding to depth $d + 1$ does not produce any new evidence variables. Let $l_d(q)$ and $u_d(q)$ denote the lower and upper bounds produced by LFI expanded to depth $d$. Then $l_{d+1}(q) \geq l_d(q)$ and $u_{d+1}(q) \leq u_d(q)$.*

For finite models, at some depth $d$ all variables will be expanded, and the bounds will be equal to the true probability. Therefore, the true probability lies between the bounds at every depth for finite models. For infinite models, the bounds do not necessarily converge. For example, consider the program:

```
def f() = Apply(f(), (x: Boolean) => x)
val query = f()
```

This program defines an infinite chain such that each variable is equal to its predecessor. The bounds at any depth will be (0, 1).

## 6. EXPERIMENTATION

We have produced two initial implementations of the LFI algorithm in Figaro, using VE and BP. We conducted two sets of experiments: the first with an infinite time hidden Markov model (HMM) and the second with unbounded and infinite probabilistic context-free grammars (PCFGs).

### 6.1 INFINITE HMM

Our first experiments were with an infinite time hidden Markov model (HMM). The model has 15 states numbered 0 to 14. At each time step, the state transitions to an adjacent state upwards or downwards, unless it is an endpoint, in which case the state stays the same. The transition probabilities are designed so that in middle states, the probability of transitioning upward or downward are approximately equal, while near the endpoints, the state will tend to transition towards the nearby endpoint. Specifically, in state $s$, where $1 \leq s \leq 13$, the state transitions to state $s + 1$ with probability $s / 14$. The effect of this is that when the state is near the middle, there is significant uncertainty about which endpoint will be reached, while closer to the endpoints there is less uncertainty.

Each state produces a Boolean emission. The emission probabilities are designed so that there is ambiguity near the middle and less ambiguity near the endpoints. Specifically, a state $s$ produces the value true with probability $s / 14$. In our experiments, the task was to determine whether state 0 or state 14 would be reached first. The amount of time until an endpoint is reached is unbounded, but reaches an endpoint with probability 1. The inference algorithm was provided with the first 10 emissions as evidence and inferred the probability that state 14 would be reached first.

Figure 2(a) shows the results of lazy VE on this problem. The figure shows the runtime of inference (blue) and the gap between the lower and upper bounds produced by the lazy algorithm (red), as a function of the depth of inference. From the figure, we see that at depth 40, the bounds gap is negligible (0.0072), while the runtime is less than 1 second. In contrast, MCMC has error of 0.047 after running for about an hour. Importance sampling failed to terminate on this problem.

### 6.2 UNBOUNDED AND INFINITE PCFGS

Our second set of experiments is with probabilistic context free grammars (PCFGs). While natural language algorithms can find the probability of specific finite strings, substring queries are more difficult as arbitrarily long and even infinitely long strings may need to be examined to determine if the substring is present. Non-lazy factored algorithms cannot answer these queries and sampling algorithms can infinitely expand. We tested our LFI algorithm on both unbounded PCFGs that produce finite strings with probability 1 and on PCFGs that produce infinite strings with positive probability.

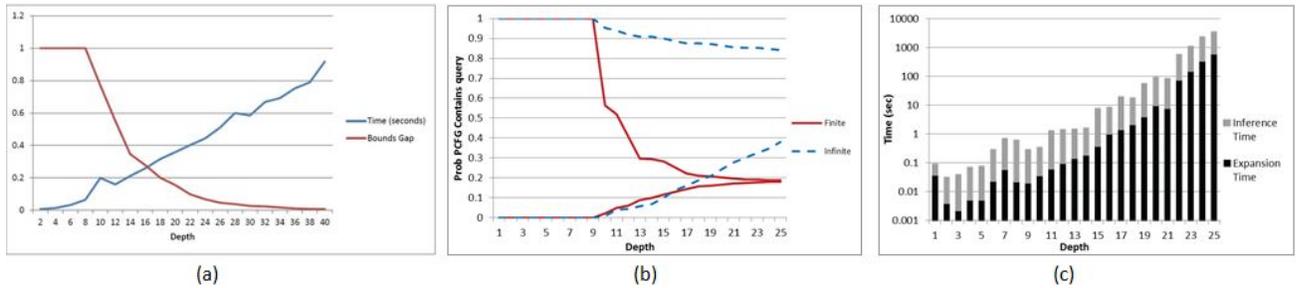

Figure 2: LFI Performance. (a) Bounds gap and runtime on infinite HMM; (b) Computed bounds on unbounded and infinite grammar; (c) runtime on infinite grammar as a function of depth

We constructed a simple PCFG with three non-terminals, where the only difference between the unbounded but finite and infinite grammars is the production probabilities. The grammar is encoded in Greibach Normal Form (GNF) (Greibach, 1965). Every production contains a terminal at the beginning. Since this expansion is recursive, expanding to a fixed depth in the LFI algorithm will bound the length of the possible strings that can be generated, and thus produce bounds on the probability of the query string.

We represented the PCFG similar to the lazy list described in the main text of the paper. As before, we define classes to generate strings. First, we define the Sequence class, which contains three sub-classes: Empty, Cons and Cont. Empty and Cons are defined the same as the lazy list, with an element representing the next terminal and an element representing the remaining non-terminals. As we are representing the PCFG in Greibach Normal Form, we added another class to more easily represent GNF expansions. The Cont class contains an element representing the next terminal, an element named next representing the left-most non-terminal, then finally an element tail representing all remaining non-terminals. This enables the lazy expansion of the grammar. That is, when a Cont is encountered in the expansion, we continually expand the next until a terminal is reached, at which point we then start expanding tail.

Figure 2(b) shows the results of a query for the probability that a string produced by the PCFG contains the substring "de", given the observation that the string contains the substring "a". We show the results for both the finite and infinite versions of the grammar, expanding using LFI to a target depth ranging from $d = 1$ to $d = 25$ using VE. The probability bounds on the queries in both grammars start to converge, and in the case of the finite grammar, do so quickly (at a depth of 21).

In Figure 2(c) we also show the running times of the model expansion and inference for the infinite grammar. The running times are dominated by the factored inference, not the lazy expansion (the graph is logarithmic). We also queried the infinite PCFG using MH and importance sampling. After 100 seconds of sampling via MH, the estimated probability of the query string is 0.001, well outside the bounds computed by lazy VE (depth of 21). MH also created more than 5 million elements and eventually ran out of memory. The importance sampler did not even produce an estimate, as once the sampler generates a world with infinite expansion, it will never terminate.

## 7. DISCUSSION AND RELATED WORK

### 7.1 CONNECTION TO LOGICAL PROBABILISTIC PROGRAMMING

Some similar ideas to the ones in this paper have appeared in the literature on logical probabilistic programming languages. Poole introduced the main ideas for probabilistic Horn abduction (PHA) (Poole, 1993). PHA inference is performed by searching for explanations of the query. An explanation is an assignment of values to probabilistic variables that entails the query. Poole's algorithm maintains a priority queue of partial explanations such that only partial explanations consistent with the query are maintained in the queue. The algorithm can terminate at any time, with a lower bound equal to the total probability of complete explanations generated, and an upper bound equal to the probability of complete explanations plus the probability of partial explanations still in the queue. A similar idea underlies inference in ProbLog (Kimmig et al., 2011). In addition, the ProbLog algorithm has been extended to automatically generate near-optimal lower bounds based on a fixed number of proofs (Renkens, Van den Broeck, and Nijssen, 2012). Finally, these ideas have also been applied by compiling the search for explanations into weighted partial MAX-SAT (Renkens et al., 2014).

The connection to our work is that an explanation in the logical setting corresponds to an assignment of values in the factored setting that is sufficient to determine an entry in the sum-of-products expression defined by the variables. Just as not all variables need to appear in an explanation, not all variables need to be assigned a value to determine the sum-of-products entry. This is one of the main ideas exploited in our algorithm.

However, there are two major differences between that literature and our work. First, those works were developed in the context of logic programming and made heavy use of logic programming tools such as proofs and SLD-resolution. These tools are not available in a functional framework, which makes developing lazy algorithms for the functional framework significantly harder.

The second, more substantial, difference is in the unit of computation. The unit of computation in the logical approaches is an explanation, which is an assignment of values to a set of probabilistic variables. There is no requirement that the explanations generated for a set of variables cover all the values of those variables. This fact resulted in the PHA-style algorithms having very weak upper bounds, although this was addressed in (Renkens et al., 2014) by allowing negation.

In contrast, in a factored framework, the unit of computation is a variable together with all of its values. Factored algorithms require that all the values of a variable be enumerated. This enables algorithms such as junction tree and VE to work. As a result, to make lazy inference work in a factored framework, we had to introduce the special value *, and ensure that * was used consistently as if it was a real value.

Connections between logical and functional probabilistic programming frameworks are still under-explored. We hope that through this work we can add to the understanding of the relationship between the fields.

### 7.2 OTHER RELATED WORK

Lazy evaluation and execution has a long history in computation, particularly as a feature of functional programming. For example, lazy evaluation of game trees using the alpha-beta algorithm allows computation of a potentially infinite search space (Hughes, 1989). In (Kiselyov and Shan, 2009), a domain specific language for probabilistic programming is embedded in OCaml, using continuations to represent a stochastic computation as a lazy search tree. The tree is traversed depth first and the probabilities of query values accumulated in a table used by the inference algorithms. IBAL (Pfeffer, 2007) provides an algorithm for solving infinite probabilistic models in PP with finite observations and also makes use of laziness to evaluate queries on infinitely large models. However, IBAL's approach only works if the evidence guarantees that only a finite part of the model can be constructed, working in a manner similar to natural language algorithms on finite strings. In (Pfeffer and Koller, 2000), the authors propose a scheme for inference with recursive probabilistic models, but it is not computationally expressed. None of these approaches use the structure of the model to determine the relevance of unexpanded variables and provide bounds on queries. Finally, (Madsen, 2004) and (Cano, Moral, and Salmerón, 2002) use laziness to simplify the computation, but nevertheless take into account all the relevant parts of the network. In contrast, the LFI approach deliberately ignores parts of the model, while characterizing precisely the effect of these parts to provide bounds on the probability of the query.

There is also a body of work related to achieving more efficient inference in Bayesian networks by exploiting the structure of the graphical models to prune irrelevant nodes and manipulate the possible factorizations (Zhang and Poole, 1994; Pearl, 1988; Shachter, 1988; Baker and Boult, 1990). These approaches all work by considering the structure of the graph and the associated dependencies. The first step of our algorithm, which expands the model, extends and builds on these concepts with a lazy expansion. Step 1 also extends ideas from knowledge-based model construction (Goldman and Charniak, 1990; Ngo and Haddawy, 1996; Haddawy, 1994). The major difference is that we do not expand until termination, instead leaving part of the computation unexpanded. This also necessitates backtracking to ensure the consistency of the expanded part of the network.

Our work is also related to box propagation methods for providing bounds to BP algorithms (Mooij and Kappen, 2008; Ihler, 2012). Anytime lifted belief propagation (de Salvo Braz et al., 2009) extends these ideas to a first-order setting. While this produces similar computations to ours, our use of * enables us to fold the calculations into the structure of the program, particularly using the compilation process described in Section 4.2. It is not obvious how to achieve the same effect using box propagation.

## 8. CONCLUSION AND FUTURE WORK

In this paper, we have presented an algorithm for LFI in PP, making factored inference a viable framework for full-fledged PP. LFI leverages the fact that not all variables in a probabilistic model are relevant to a particular query and provides bounds on the query probability by only exploring the most relevant portions of the model. We have provided a basic algorithm, and several optimizations to improve efficiency and accuracy. Experimental results using an implementation of LFI in Figaro demonstrate the potential of this approach for providing tractable, factored algorithms for PP.

Our main goal for future work is to develop and analyze further a range of specific lazy factored algorithms. In particular, we would like to investigate the interaction between BP's approximate computations and the lazy bounds. In addition, we plan to investigate intelligent expansion strategies that expand portions of the program that are most likely to improve the answer to the query.